%% file: main.tex
\documentclass[conference]{IEEEtran}
\IEEEoverridecommandlockouts

\usepackage{hyperref}
\usepackage{fancyhdr}
\usepackage[ruled,vlined,linesnumbered]{algorithm2e}

\SetCommentSty{mycommfont}

\usepackage{multirow}

\usepackage{makecell}
\usepackage{color,soul}

\usepackage{pifont}
\usepackage{amssymb}

\usepackage{diagbox}
\usepackage{listings}

\usepackage{tikz}
\usepackage{amsmath}
\usepackage{amsfonts}

\usepackage{filecontents}
\usepackage{ctable}

\usepackage{cite}
\usepackage{amsmath,amssymb,amsfonts}
\usepackage{algorithmic}
\usepackage{graphicx}
\usepackage{textcomp}
\usepackage{xcolor}
\def\BibTeX{{\rm B\kern-.05em{\sc i\kern-.025em b}\kern-.08em
    T\kern-.1667em\lower.7ex\hbox{E}\kern-.125emX}}
    
\usepackage{color}
\definecolor{codegreen}{rgb}{0,0.6,0}
\definecolor{codegray}{rgb}{0.5,0.5,0.5}
\definecolor{codepurple}{rgb}{0.58,0,0.82}
\definecolor{backcolour}{rgb}{0.95,0.95,0.92}
\definecolor{textblue}{rgb}{.2,.2,.7}
\definecolor{textred}{rgb}{0.54,0,0}
\definecolor{textgreen}{rgb}{0,0.43,0}
\definecolor{codered}{rgb}{201,72,12}

\definecolor{calpolypomonagreen}{rgb}{0.12, 0.3, 0.17}
\definecolor{cobalt}{rgb}{0.0, 0.28, 0.67}
\lstdefinestyle{tt1}{
language=Python,
basicstyle=\linespread{0.98}\ttfamily\footnotesize,
breaklines=true,
numbers=left,
frame=single,
numberstyle=\tiny, 
stepnumber=1,
numbersep=5pt, 
tabsize=4,
commentstyle=\color{textred},   
keywordstyle=\bfseries\color{codegreen},
stringstyle=\color{textgreen},
columns=fullflexible,
keepspaces=true,
xleftmargin=\parindent,
showstringspaces=false,
otherkeywords = {True, False},
keywordstyle=[3]\color{codegreen}\bfseries,
keywords=[3]{__device__},
keywordstyle=[4]\color{cobalt}\bfseries,
keywords=[4]{Efficient_TT, Eff_TTEmbedding, efficient_tt, Eff_TT, Efficient_TTEmbedding, Efficient_TT_Embedding},
}

\begin{document}

\title{
 Chats-Grid: An Iterative Retrieval Q\&A Optimization Scheme Leveraging Large Model and Retrieval Enhancement Generation in smart grid
}
\author{Yunfeng Li,~\IEEEmembership{Student Member,~IEEE,}
Jiqun Zhang,~\IEEEmembership{Student Member,~IEEE,}
Guofu Liao,~\IEEEmembership{Student Member,~IEEE,}
\\ Xue Shi,~\IEEEmembership{Student Member,~IEEE,}
Junhong Liu,~\IEEEmembership{Member,~IEEE,}

\thanks{Yunfeng Li, Jiqun Zhang, and Guofu Liao are with the Department of Electronics and Information Engineering, Shenzhen University, China (liyunfeng2019@email.szu.edu.cn; zjq@email.szu.edu.cn; lgf@email.szu.edu.cn).(Corresponding author: Jiqun Zhang).}
\thanks{Junhong Liu is with the Department of Electrical and Electronic Engineering,
The University of Hong Kong, Hong Kong SAR, China (email:
jhliu@eee.hku.hk).}}

\maketitle

\thispagestyle{fancy}
\lhead{}
\rhead{}
\chead{}
\rfoot{}
\cfoot{}
\renewcommand{\headrulewidth}{0pt}
\renewcommand{\footrulewidth}{0pt}

\input{sections/01_abstract}


\input{sections/02_intro}


\input{sections/04_TT_core}

\input{sections/05_Reordering}

\input{sections/06_Pipeline}
\input{sections/07_evaluation}

\input{sections/09_related_work}
\input{sections/10_conclusion}

\input{sections/11_acknowledgment}

\bibliographystyle{./IEEEtran}
\bibliography{ref}


\end{document}

%% file: sections/01_abstract.tex
\begin{abstract}
With rapid advancements in artificial intelligence, question-answering (Q\&A) systems have become essential in intelligent search engines, virtual assistants, and customer service platforms. However, in dynamic domains like smart grids, conventional retrieval-augmented generation(RAG) Q\&A systems face challenges such as inadequate retrieval quality, irrelevant responses, and inefficiencies in handling large-scale, real-time data streams.
This paper proposes an optimized iterative retrieval-based Q\&A framework called Chats-Grid tailored for smart grid environments. In the pre-retrieval phase, Chats-Grid advanced query expansion ensures comprehensive coverage of diverse data sources, including sensor readings, meter records, and control system parameters. During retrieval, Best Matching 25(BM25) sparse retrieval and BAAI General Embedding(BGE) dense retrieval in Chats-Grid are combined to process vast, heterogeneous datasets effectively. Post-retrieval, a fine-tuned large language model uses prompt engineering to assess relevance, filter irrelevant results, and reorder documents based on contextual accuracy. The model further generates precise, context-aware answers, adhering to quality criteria and employing a self-checking mechanism for enhanced reliability.
Experimental results demonstrate Chats-Grid’s superiority over state-of-the-art methods in fidelity, contextual recall, relevance, and accuracy by 2.37\%, 2.19\%, and 3.58\% respectively. This framework advances smart grid management by improving decision-making and user interactions, fostering resilient and adaptive smart grid infrastructures.
\end{abstract}

%% file: sections/02_intro.tex
\section{Introduction}
The advent of smart grid technologies has revolutionized the energy sector by integrating advanced metering, real-time monitoring, and two-way communication systems. These advancements have enabled more efficient energy management, greater resilience to disruptions, and an enhanced consumer experience. However, the growing complexity of smart grid systems necessitates innovative solutions for real-time decision-making, operational optimization, and effective consumer interaction. One such promising avenue is the integration of intelligent question-and-answer (Q\&A) systems, leveraging large language models (LLM) and advanced retrieval techniques to address the dynamic demands of smart grids. These systems can support grid operators by providing accurate, context-aware responses to operational queries and enhancing user engagement through natural language interactions.

Existing research has explored various facets of artificial intelligence (AI) applications in smart grids, including demand forecasting, fault detection, and grid stability assessment~\cite{omitaomu2021ai}. Notably, hybrid frameworks that combine machine learning and knowledge-based approaches have shown significant promise in decision-making and emergency response scenarios~\cite{glukhikh2022cbr}. The development of retrieval-enhanced generation models has further advanced Q\&A systems by improving the accuracy and contextual relevance of responses~\cite{razzak2024deep}. Despite these advancements, challenges such as scalability, real-time performance, and adaptability to evolving grid conditions remain inadequately addressed.

With the development of AI technologies,
question-answering (Q\&A) systems have played a crucial role
in fields such as information retrieval, knowledge management,
and human-computer interaction. 
Traditional Q\&A systems typically rely on a series of structured steps to process utilizer queries, including rule-based methods and retrieval-based methods. However, recent advancements in computing power have accelerated development of LLM technologies, making generative LLM-based Q\&A systems increasingly prevalent. Since OpenAI released the GPT-3.5 model in 2022, its exceptional performance in general domains has demonstrated the vast potential of LLM. Subsequently, major companies such as Meta, Google, and Baidu, have also launched their own LLM (e.g., LLama~\cite{touvron2023llama}, Gemini~\cite{team2023gemini}, Ernie Bot~\cite{ren2023evaluation}), further advancing the field.

Although generative Q\&A systems can flexibly generate responses, they still face challenges in terms of accuracy and reliability. For instance, models could produce fabricated information on certain specific questions, a phenomenon known as ``hallucination"~\cite{huang2023survey}. To improve response accuracy, retrieval-augmented generation (RAG)~\cite{lewis2020retrieval} Q\&A systems synthesize generative LLM with external retrieval resources, retrieving relevant background knowledge from these resources and providing it as supplementary input to the generative LLM. This approach helps the generative LLM respond based on authentic external information, thereby effectively reducing the ``hallucination" phenomenon of the model. In this context, RAG technology has gradually become a key technology for applications
such as search engines, chat engines, and agents, improving
the accuracy and practicality of model responses by combining
retrieval with generation. 

Traditional RAG Q\&A systems are typically involve three steps: building a text index, retrieving text content, and generating responses. However, this method only relies on a single retrieval to find relevant texts and does not further optimize the retrieval and generation processes, thus limiting the system's retrieval and response quality. As RAG technology advances, enhanced and modular retrieval augmentations have emerged, but they still face the following issues:

\textbf{1) Insufficient retrieval quality:} This includes poor relevance between the user queries and the retrieved content, incomplete retrieval results, and excessive verbosity due to the large and heterogeneous datasets for smart grid environments.

\textbf{2) Limited response quality:} Due to the impact of retrieval quality, where LLM could be misled by irrelevant contexts. Additionally, the confusion between parametric and non-parametric memories can also lead to suboptimal performance of LLM, which further compromises the querying quality for smart grid environments.

Current RAG optimization research
has focused on three areas respectively:

\textbf{1) Optimization based on model training and fine-tuning:} These methods focus on introducing optimization techniques during the pre-training and fine-tuning stages of retrieval and generative models to improve their overall performance.  Yu~\cite{yu2023augmentation} fine-tuned the retriever using feedback signals to align it with the preferences of the LLM, enhancing their coordination. Cheng~\cite{cheng2024lift} fine-tuned the generator to adapt the LLM to the input structure of text pairs. The Retro~\cite{borgeaud2022improving} method pre-trains the generative model from scratch, encodes retrieved documents using a Transformer encoder, and integrates them into the token hidden states of the attention layer using a cross-attention mechanism, achieving deep integration between retrieved information and the generative model. This approach not only reduces the model's parameter size but also achieves superior performance in perplexity metrics, demonstrating its potential in enhancing generation quality.

\textbf{2) Optimization based on data sources:} These methods extend the retrieval of single unstructured texts in knowledge bases to using structured data or data generated by the model itself to further improve the effectiveness of RAG. For instance, Ret-LLM~\cite{modarressi2023ret} constructs a knowledge graph as memory using past Q\&A data for subsequent dialogue reference. Similar to RET-LLM, the Selfmem method creates an unlimited memory pool, adding each round of LLM output to the pool and selecting the most appropriate memory through a memory selector to improve subsequent generation. The Surge~\cite{kang2023knowledge} method retrieves relevant subgraphs from knowledge graphs to improve model response quality, ensuring the relevance of retrieved subgraphs to questions through perturbed word embeddings and contrastive learning. Luyu Gao et al. proposed the Hypothetical Document Embedding (HyDE) method~\cite{gao2022precise}. When a utilizer provides a query, the HyDE method, unlike traditional retrieval approaches that rely on query-based search, first hypothesizes an answer and then retrieves relevant information from the knowledge base based on this hypothetical answer. It focutilizes on the similarity between the hypothesized answer and the actual answer rather than seeking embedding similarity for the query, thereby improving retrieval recall.

\textbf{3) Optimization based on the retrieval process:} These methods improve the traditional single-retrieval mode of RAG by introducing multiple retrieval strategies such as iterative retrieval, recursive retrieval, and adaptive retrieval~\cite{jiang2023active} to enhance retrieval performance and effectiveness. Iterative retrieval allows the model to perform multiple rounds of retrieval, enhancing the depth and relevance of the information obtained. Recursive retrieval refers to using the results of the previous retrieval as input for subsequent retrievals, which helps deeply mine relevant information, especially when dealing with complex or multi-step queries. Ori Ram~\cite{ram2023context} proposed a method called In-Context RALM, which generates a small number of incomplete answers through the LLM each time, then retrieves texts similar to these answers and concatenates them into the prompt for subsequent answer generation, completing the answer through multiple retrieval processes. Akari Asai et al. introduced a method called Self-RAG~\cite{asai2023self}, which marks utilizer queries and retrieved content using two key markers: retrieval markers and criticism markers. Retrieval markers determine the need to invoke the retrieval model, while criticism markers assess the relevance of retrieved content to the query. Finally, the results are re-ranked based on relevance scores to optimize the model's output. Zhangyin Feng~\cite{feng2024retrieval} proposed an Iterative Retrieval Generation Collaboration (ITRG) framework, which enhances response quality through multiple iterations. It initially retrieves information using the original query, and each subsequent iteration involves two steps: 1) generating a response with potential hallucinations using the LLM and retrieved content; 2) retrieving relevant content using the model's response from the previous step.

\begin{figure} [h] \small
    \centering
    \includegraphics[width=0.9\linewidth]{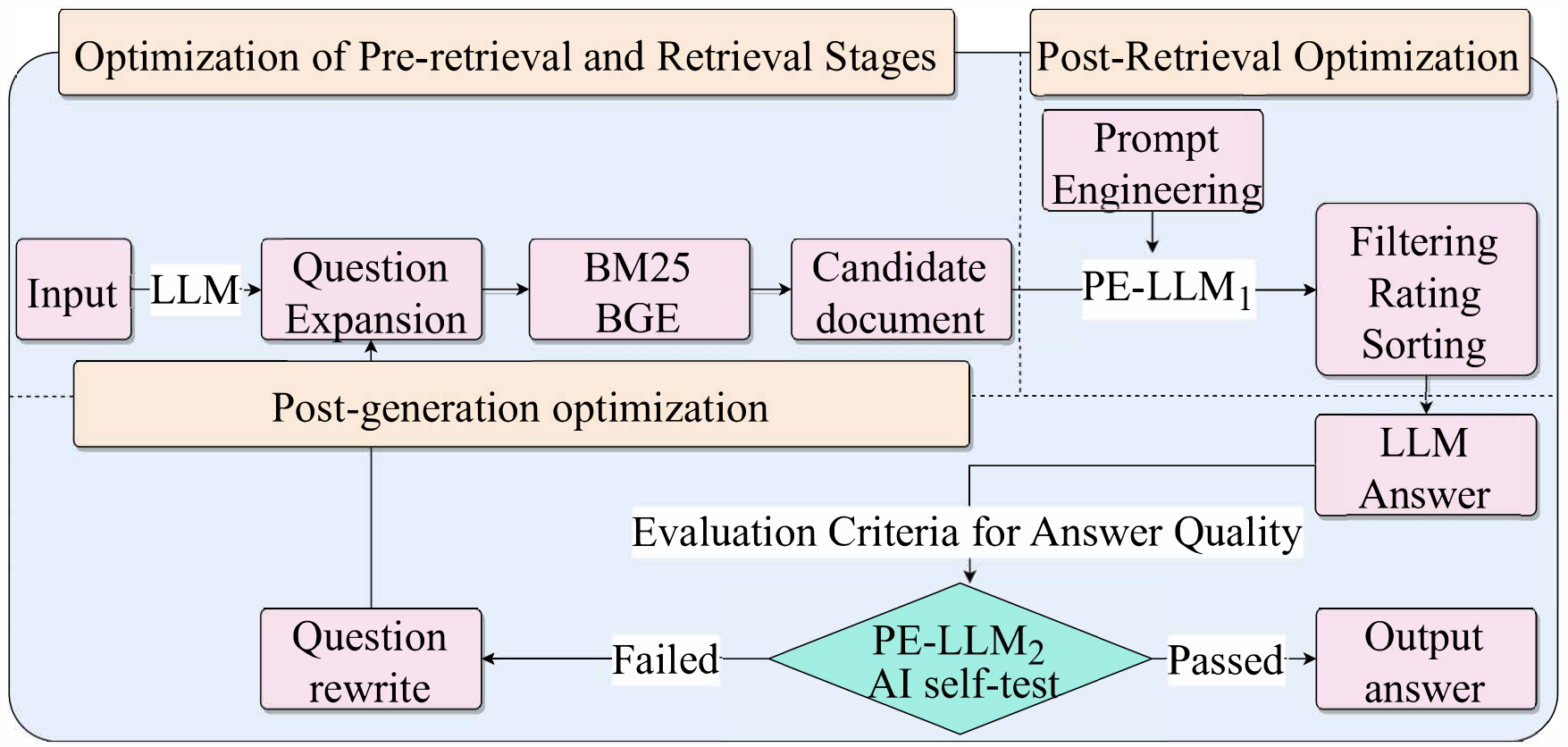}
    \caption{Optimized RAG System Process.}
    \label{fig:fig1}
\end{figure}

Research published by Lexin Zhou~\cite{zhou2024larger} in the authoritative journal Nature pointed out that subtle adjustments in prompt engineering can have profound impacts on the output of LLM, emphasizing the need for a rigorous and meticulous approach in designing prompts to ensure output stability and reliability. Although RAG technologies have made significant progress, there is a lack of comprehensive Q\&A system architecture optimized for each critical stage: pre-retrieval, retrieval, post-retrieval processing, and response generation. Finally, this paper introduces ``Chats-Grid", an iterative retrieval Q\&A optimization scheme tailored to smart grid environments, featuring stage-specific optimizations and refined strategies to construct a comprehensive solution. By leveraging large models and retrieval enhancement generation, Chats-Grid aims to provide a robust framework for optimizing grid operations and consumer interactions. The approach addresses key limitations in existing systems, including response latency, contextual accuracy, and adaptability, thus paving the way for a more resilient and efficient smart grid infrastructure. In this study, we evaluate the proposed methodology's effectiveness through comprehensive simulations and case studies, highlighting its potential to transform smart grid management and user engagement. The overall optimization framework is detailed in Figure~\ref{fig:fig1}. Overall, we make the following contributions in this paper:
\begin{itemize}
    \item \textbf{Before retrieval}, we expand the query to increase the depth and scope of retrieval; during retrieval, we enhance retrieval robustness by employing both dense and sparse retrieval methods concurrently.

\item \textbf{After retrieval}, we fine-tune the LLM using prompt engineering to score the relevance of retrieved candidate documents, filtering out irrelevant documents for the smart grid environments to the following contributions and reordering them.

\item \textbf{During the generation stage}, we further avoid hallucinations through an answer self-checking mechanism and evaluate response quality using a quintuple assessment standard. If the self-check fails, the query is rephrased and re-entered into the Chats-Grid system for processing.

\item \textbf{Comprehensive Q\&A system architecture optimization}, we are pioneers in applying comprehensive Q\&A system architecture to improve efficiency and  answer quality in smart grid. Specifically, Chats-Grid shows improvements of 2.37\%, 2.19\%, and 3.58\% in fidelity, context recall rate, and answer accuracy over Self-RAG, respectively, and 0.94\%, 4.39\%, and 2.45\% over ITRG.
\end{itemize}

The remainder of this paper is organized as follows. Section II presents the preliminaries of the proposed method. Section III offers a comprehensive overview of the design aspects of the Eff-TT table and sorting indexes methods aimed at enhancing the performance of Eff-TT tables. Section IV offers an overview of the TT-based pipeline and the training system design, along with its solution for addressing Read-after-write conflicts within the DLRM training pipeline. Case studies are conducted in Section V and Section VI concludes this paper.


        



%% file: sections/04_TT_core.tex
\section{Preliminaries}
The construction of current mainstream Q\&A systems typically follows a process divided into three stages: index construction, pre-search and retrieval, and post-search and generation. In the index construction phase, text segmentation strategies are employed to address challenges related to natural text redundancy and model input limitations. During the pre-search stage, the optimization of user queries is performed to improve retrieval accuracy, achieved through the use of LLM and prompt engineering. In the retrieval phase, a search engine is utilized to identify similar texts. The post-retrieval stage involves text filtering and reordering, streamlining the context to enhance LLM attention. Finally, a large language model is selected during the generation phase to produce accurate and user-friendly answers.

\subsection{Index Construction}
The core task of the indexing phase is to balance information integrity with retrieval performance. Since natural language texts are often lengthy and may exceed the maximum input sequence length of Transformer models, it is essential to segment long texts to fit model constraints and reduce system overhead. An effective segmentation strategy should preserve sufficient contextual information, minimize noise, and enhance retrieval efficiency. Common text segmentation techniques include sliding windows, merging, and block summarization. For example, sentence-window retrieval improves query precision by dividing the text into shorter sentences and expanding the context window around retrieval results to enhance contextual understanding. Additionally, segmentation strategies may involve optimizing block organization structures, such as building hierarchical indexes or knowledge graphs, to accelerate data retrieval and processing. One such method, automatic merging retrieval, uses a parent-child node structure, replacing multiple related child nodes with their parent node during retrieval, thereby improving both accuracy and efficiency.

\subsection{Pre-search stage and retrieval stage}\label{tt_intro}
The primary goal of the pre-retrieval phase is to optimize the user's initial query to improve the accuracy and relevance of retrieval results. These optimizations are typically achieved through the use of LLM and prompt engineering. Since user queries often suffer from unclear descriptions, illogical structures, or inappropriate framing, this phase focuses on addressing issues such as poorly worded queries, linguistic complexity, and ambiguity.

The retrieval phase aims to identify similar texts within the knowledge base based on the user's query, a task performed by retrieval systems. These systems are typically categorized into sparse retrievers and dense retrievers. Sparse retrievers, such as Term Frequency-Inverse Document Frequency(TF-IDF) and Best Matching 25(BM25) algorithms, are considered earlier-generation methods but remain widely used in various fields due to their efficient encoding and stability. Dense retrievers, on the other hand, utilize deep learning techniques, employing neural networks to learn dense vector representations of documents and queries. Common models include the BGE series, text-embedding-ada-002 and Moka Massive Mixed Embedding(M3E). Current mainstream approaches involve transforming text into embedding vectors using these models and then calculating cosine similarity between vectors~\cite{karpukhin2020dense} to measure textual similarity. The embedding process can be expressed as equation~\ref{1}:
\begin{equation} \label{1}
\vec{q}=\operatorname{Encoder}_q(q) ; \overrightarrow{d_l}=\operatorname{Encoder}_d\left(d_i\right)
\end{equation}

where, ${Encoder}_q$ and ${Encoder}_d$ are embedded vector models, which usually share weights or architectures for mapping textual data into a vector space~\cite{cuconasu2024power}. This mapping allows the text data to be represented as dense vectors of fixed length, capturing the semantic relationships between words, which facilitates subsequent computation and processing. After obtaining the vector representation of the text, similar texts can be identified by calculating the cosine similarity, as demonstrated in equation~\ref{2}.
\begin{equation} \label{2}
\text { Similarity }=\cos \theta=\frac{\vec{q} \cdot \overrightarrow{d_l}}{\|\vec{q}\|\|\overrightarrow{d_{\imath}}\|}
\end{equation}

where $\vec{q} \cdot \overrightarrow{d_l}$ is the dot product of vectors $\vec{q}$ and $\overrightarrow{d_l}$, ${\|\vec{q}\|}$ and $\|\overrightarrow{d_{\imath}}\|$ are the lengths of ${\vec{q}}$ and $\overrightarrow{d_{\imath}}$ . The value of cosine similarity ranges from -1 to 1, with values closer to 1 indicating higher similarity and values closer to -1 indicating lower similarity. When selecting an embedding vector model, three key considerations must be addressed: the efficiency of the retrieval, the quality of the embedding vectors, and the alignment between the task, the data, and the model.

In addition to selecting an appropriate embedding vector model, model fine-tuning is an effective approach to enhance retrieval performance, particularly in highly specialized domains~\cite{houlsby2019parameter}. Common fine-tuning methods include Supervised Fine-Tuning(SFT), LLM-Supervised Retrieval(LSR), and Adapter~\cite{jiang2023llmlingua}. SFT, akin to traditional fine-tuning techniques, involves constructing a fine-tuning dataset based on domain-specific data for training. The LSR method leverages the output of a LLM to supervise and fine-tune the retrieval model. When the retriever is hosted on a cloud service platform or when cost-efficiency is prioritized, fine-tuning can be achieved through the Adapter module.

\subsection{Post-retrieval and Generation Stages}

Retrieved texts often contain low-relevance or erroneous information, and excessive context can introduce noise, hindering LLM from capturing key information and potentially leading to omissions. To mitigate these issues, the post-retrieval phase employs techniques such as text filtering and re-ranking. Text filtering streamlines the context, reduces LLM resource consumption, and minimizes response latency. For example, the LLMLingua method~\cite{liu2024lost} uses small language models to detect and remove irrelevant markers.

Re-ranking improves the LLM's focus on relevant documents and can be classified into rule-based and model-based approaches. Rule-based methods include strategies such as diversity-based or relevance-based re-ranking, while model-based approaches employ advanced tools, such as the bge-reranker-large model developed by the Beijing Academy of AI, to optimize result quality.

In the generation phase, selecting an appropriate LLM for answer generation is crucial. Options include proprietary models such as GPT-3.5, which offer high concurrency, strong performance, and no server maintenance burden, but also present data privacy risks and lack fine-tuning capabilities. Alternatively, locally deployed open-source models like Llama or Chat Generative Language Model(ChatGLM) offer greater flexibility and security, though they require significant computational resources. To enhance answer quality, larger parameter models or fine-tuning techniques can be used to incorporate domain-specific knowledge or adapt to particular data formats and styles.


%% file: sections/05_Reordering.tex
\section{Optimization of Q\&A Framework for Iterative  Question Answering}


Figure 1 presents the overall process of the iterative retrieval question-answering optimization scheme, which mainly consists of three parts: pre-retrieval and retrieval stage optimization, post-retrieval optimization, and post-generation optimization. Specifically, the input information is first processed by a LLM to perform question expansion. Subsequently, the BM25 and BGE algorithms are used to retrieve candidate documents. On this basis, we use prompt engineering(PE) to fine-tune the LLM to form a scoring-capable $PE-LLM_1$. This model is used to filter, score, and rank the candidate documents. Through the above steps, the expanded questions and candidate documents can be obtained and input into the LLM to get corresponding answers. Then, according to the answer quality evaluation criteria we formulated, we use prompt engineering to fine-tune the LLM again to form a $PE-LLM_2$ with the ability to evaluate answer quality. If the self-test of $PE-LLM_2$ passes, the answer is output; if it fails, the question is expanded and the process restarts. This process comprehensively covers all links of the question-answering system from input to output, aiming to effectively improve the quality and effect of information retrieval and answer generation.

\subsection{Optimization of Pre-retrieval and Retrieval Stages}
\label{sec: eff_tt}
\begin{figure} [h] 
    \centering
    \includegraphics[width=0.9\linewidth]{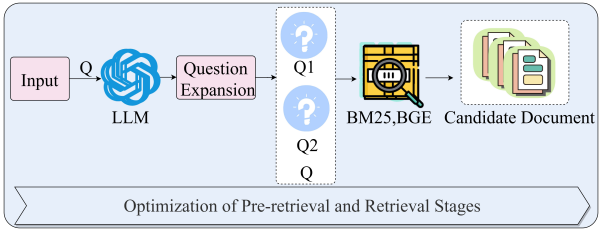}
    \caption{Query Expansion from Original Query.}
    \label{fig:fig2}
\end{figure}

User queries often suffer from issues such as vague descriptions, unclear logic, or incorrect framing, which negatively affect retrieval quality. To address these challenges, this study employs LLM to rephrase and expand the original queries text before retrieval. This approach enhances the scope and depth of document retrieval by improving query clarity and structure. As depicted in Figure ~\ref{fig:fig2}, we show the process of the original queries text expansion.

Figure 2 presents the process architecture of the pre-retrieval and retrieval stage optimization. First, the user inputs the question "Q", and then the LLM is used to perform question expansion. Experimental verification shows that the optimal effect is achieved when the original query "Q" is expanded into "Q1" and "Q2". These expanded questions, together with the original question, are input into the hybrid retrieval module constructed by BM25 and BGE, and finally a candidate document set is obtained. This process lays a solid foundation for the subsequent information processing and retrieval links, and plays a crucial role in improving retrieval efficiency and accuracy.

\begin{figure} [h] 
    \centering
    \includegraphics[width=0.9\linewidth]{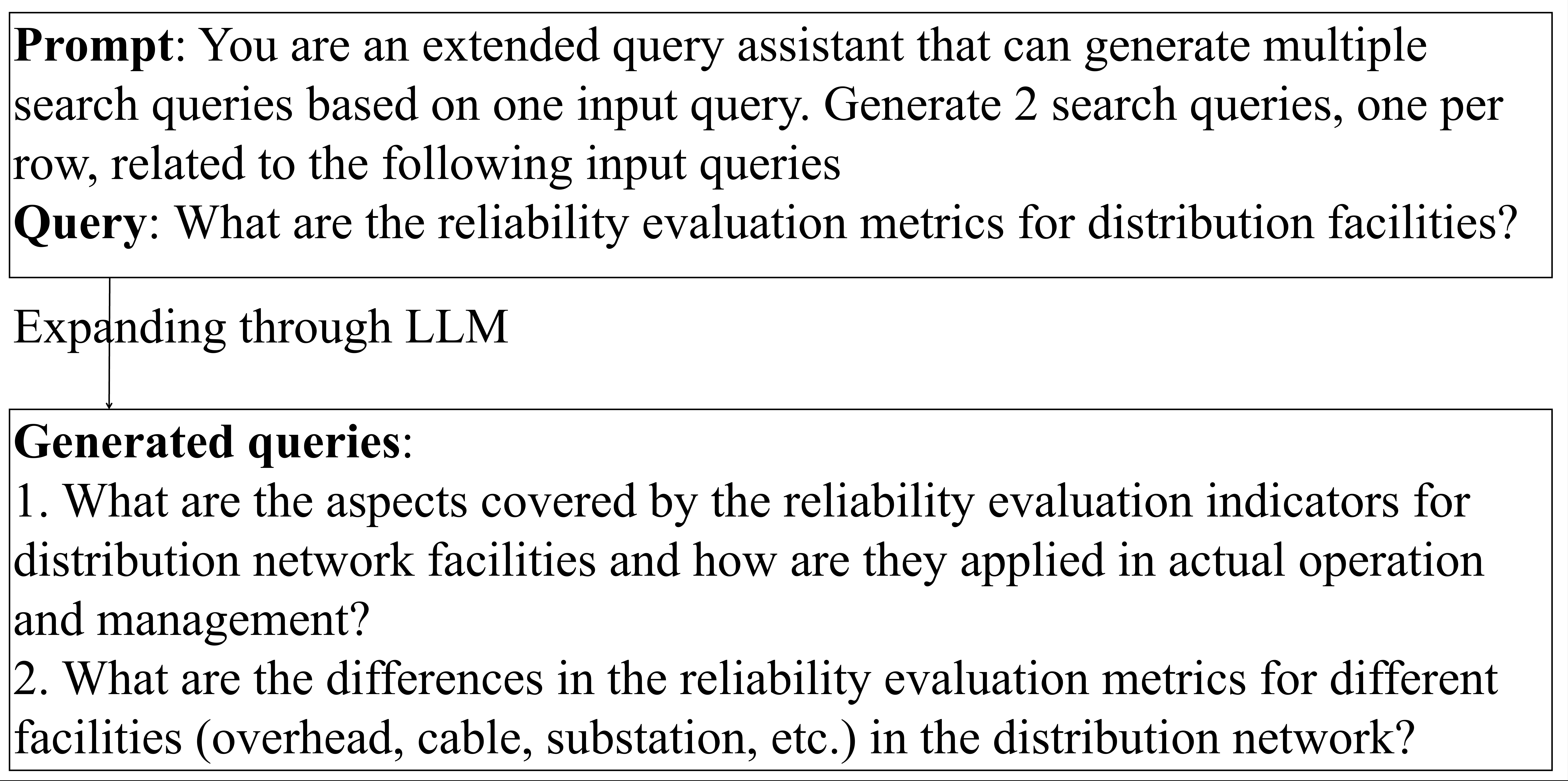}
    \caption{Query Expansion of the Original Query}
    \label{fig:fig3}
\end{figure}

Since the way users ask questions may affect the quality of retrieval, in this paper, before retrieval, the LLM is used to rewrite and expand the user's questions, so as to improve the coverage and depth of retrieved documents in the retrieval stage. The query expansion process is shown in Figure~\ref{fig:fig3}.

After completing the query expansion task, in order to efficiently mine relevant information, it is necessary to carry out parallel retrieval on multiple expanded queries simultaneously. Among many retrieval strategies, the dense retrieval method based on calculating the cosine similarity of the embedding vector model is widely used. The dense retrieval model selected in this paper is the BGE, a general vector model developed and open-sourced by the Beijing Academy of Artificial Intelligence. This model can effectively capture text semantics and provide strong support for retrieval in most scenarios. However, this dense retrieval method relying on the embedding vector model has inherent limitations. When facing emerging words or professional terms, due to the lack of training data, the BGE model is difficult to fully learn their semantic features, thus greatly reducing the accuracy of relevant document retrieval.
In view of this, this paper innovatively combines the dense retrieval of BGE with the sparse retrieval method of BM25. The BM25 algorithm can efficiently calculate the relevance between documents and queries based on term frequency and inverse document frequency, and its score can be obtained through Equation ~\ref{3} and Equation ~\ref{4}.

\begin{equation} \label{3}
\small
\operatorname{Score}(D\!,\! Q) \! = \! \sum_{i=1}^n IDF\left(q_i\right) \! \cdot 
\frac{f\left(q_i, D\right) \! \cdot \! \left(k_1 \!+ \!1\right)}{f\left(q_i \!, \! D\right) \! + \! k_1  \! \cdot \! \left(1 \! - \! b \! + \! b \! \cdot \! \frac{|D|}{a v g d l}\right)}
\end{equation}

\begin{equation} \label{4}
\operatorname{IDF}\left(q_i\right)=\log ((\mathrm{M}-\mathrm{m}+0.5) /(\mathrm{m}+0.5))
\end{equation}

Where $M$ is the total number of documents, $m$ is the number of documents containing the word, $D$ is a document, $\mathrm{Q}$ is a query statement, $q_i$ is the the word in the query statement, $f\left(q_i, D\right)$ is the word frequency of word  in document $D$, $|D|$ is the length of document $D$, and $avgdl$ is the average of the lengths of all the documents in the set of documents. $k_1$ and $b$ are the tuning parameters.

The embedding vector model can usually capture semantic information and understand the similarity between words, while the BM25 algorithm is better at precise matching based on keywords. Using the embedding vector model and the BM25 algorithm for retrieval at the same time, two sets of search results are obtained. Combining the two can make use of both semantic similarity and keyword matching degree to improve the accuracy of retrieval. Through the combination of dense retrieval and sparse retrieval, a more comprehensive information coverage is achieved, the processing ability of long-tail queries is increased, and the robustness of the query is improved.



\subsection{Flowchart of Post-retrieval Optimization}

\begin{figure}[h!]
    \centering
    \includegraphics[width=0.9\linewidth]{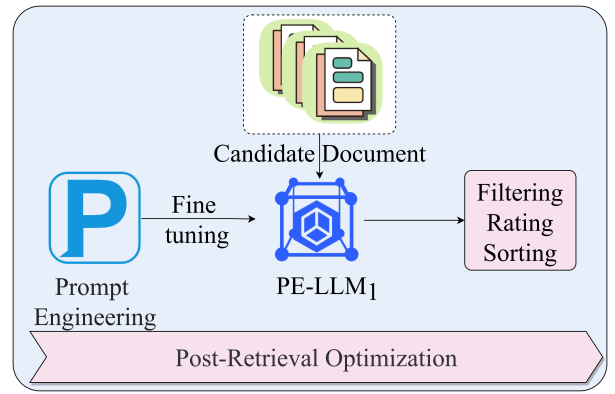}
    \vspace{-5pt}
    \caption{ Flowchart of Post-retrieval Optimization}
    \label{fig:fig4}
\end{figure}

The post-retrieval optimization process is shown in Figure~\ref{fig:fig4}. We use prompt engineering to fine-tune the LLM to obtain $PE-LLM_1$. Then, the candidate documents are input into $PE-LLM_1$, and finally, $PE-LLM_1$ performs operations such as filtering, rating, and ranking on the candidate documents to provide more accurate reference documents for the subsequent answer generation.

After obtaining relevant documents through parallel retrieval, it is necessary to carry out reordering operations on these documents. Assuming that each retrieval engine returns the top N similar texts, then after M times of parallel retrieval, N*M results will be obtained. In this case, it is necessary to use a sorting method to sort these merged documents. Among them, the Reciprocal Rank Fusion (RRF) method is more commonly used, and the documents are usually arranged in reverse order of relevance. However, for LLM, this is not the best sorting method. The middle amnesia characteristic of LLM shows that when the useful text is in different positions of the prompt words, the answering effect of the model will be better [22]. And some research has confirmed that in the vector space, those documents that are close to the query but contain useful information may have a negative effect on the model's answer [18]. In view of this, this paper uses LLM (this LLM is based on the Gpt3.5-Turbo model and has been fine-tuned using 100 artificially constructed scoring data sets) to score each retrieval result according to its helpfulness in answering the question, rather than simply marking it as Relevant or Supported, and then eliminating those documents that are of no help in answering the question. The specific scoring criteria are shown in Table ~\ref{tab:tab1}.

\setcounter{table}{0}
\begin{table}[h!]\small
 \caption{Scoring Criteria for Constructing the Rating Dataset}
    \centering
     \vspace{-5pt}
    \scalebox{0.6}{
   \begin{tabular}{ll}
\toprule[1.5pt]
Score & Scope \\
\hline 0 &  
Documents that do not contain any useful information \\
 1-3 &  
Contains only some relevant background information and does not directly provide 
an answer. \\
$4-9$ & 
Contains direct evidence of the answer to the question and is scored higher based on completeness.
\\
10 & A golden document that contains all the information needed to answer the question.\\
\bottomrule[1.5pt]
\end{tabular}}
\label{tab:tab1}
\end{table}

\begin{figure} [h!] \small
    \centering
    \includegraphics[width=0.9\linewidth]{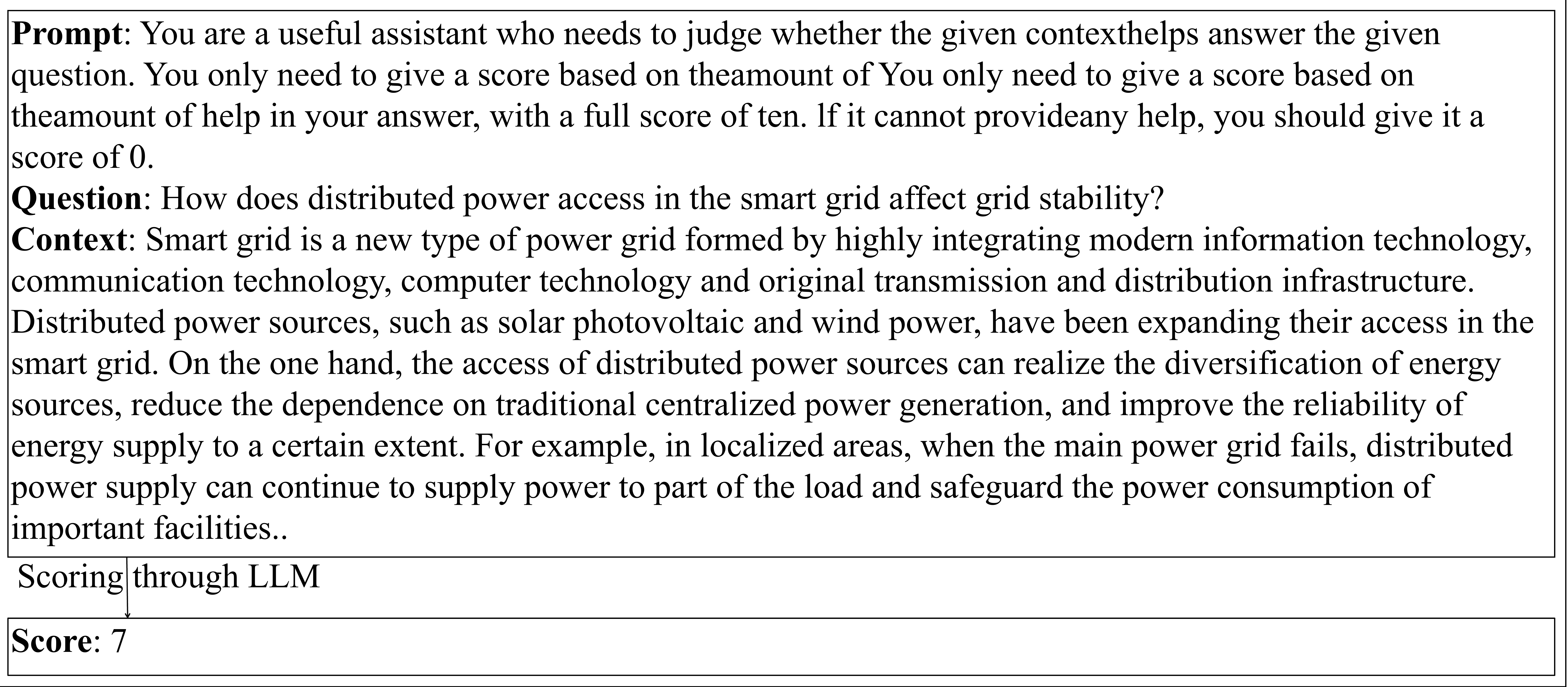}
    \caption{Scoring Document Relevance Using an LLM.}
    \label{fig:fig5}
\end{figure}

The fine-tuned large-scale model is combined with prompt words to score the candidate documents, as shown in Figure~\ref{fig:fig5}. During the model inference process, in order to maintain the consistency of the scoring, the temperature parameter of the model needs to be set to 0. After obtaining the score of each document, the system will first eliminate those documents with a score of 0, and then use these scores to sort the documents.

\subsection{Optimization in the Generation Stage}\label{bijection}

\begin{figure} [h] \small
    \centering
    \includegraphics[width=0.9\linewidth]{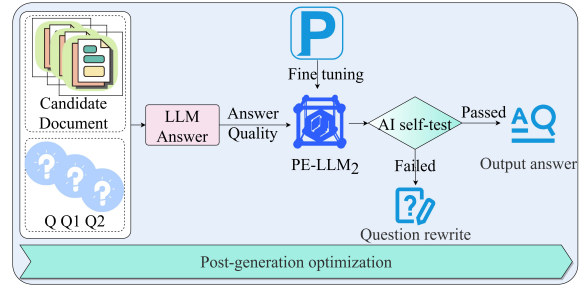}
    \vspace{-5pt}
    \caption{ Flowchart of Post-Generation Optimization}
    \label{fig:fig6}
\end{figure}

The process of post-generation optimization is shown in Figure~\ref{fig:fig6}. First, the candidate documents generated in the previous two stages and "Q, Q1, Q2" (i.e., the original question and its expanded questions) are input into the LLM to generate an answer. Subsequently, prompt engineering is used to construct $PE-LLM_2$ (a model with the ability to evaluate answer quality) to assess the quality of the generated answer. $PE-LLM_2$ will perform an AI self-check. If the self-check passes, the corresponding answer will be output; if the self-check fails, the question rewriting step will be executed, and the iterative retrieval process will be entered again to ensure the quality and accuracy of the final output answer, thereby improving the performance and reliability of the entire system in information processing and Q\&A. Next, the answer quality evaluation criteria and the AI self-check mechanism will be introduced in detail.

\textbf{Answer Quality Evaluation Criteria}

Traditional natural language processing evaluation metrics, such as Exact Match (EM), Bilingual Evaluation Understudy(BLEU), etc., mainly focus on the degree of direct text repetition. Therefore, for texts with similar semantics but different forms of expression, misjudgments are likely to occur. Different from these, Trulens introduced a triple-element evaluation criterion, comprehensively considering three dimensions: context relevance, faithfulness, and answer relevance. Ragas, on this basis, further added indicators such as answer similarity, context recall rate, and precision.
We carried out ablation experiments on the above-mentioned indicators, extracted the important indicators that play a key role in the evaluation, and based on this, proposed a five-element evaluation criterion. This criterion covers the following five dimensions:
\begin{enumerate}
    \item Faithfulness: It is used to evaluate the degree of association between the generated answer and the context, and measures whether the answer accurately reflects the context information.
    \item Context Recall Rate: Also focusing on the correlation between the generated answer and the context, it judges whether the context contains the key information sufficient to support the generated answer.
    \item Answer Relevance: Aims to evaluate the degree of close association between the generated answer and the original question, ensuring that the answer closely adheres to the question topic.
    \item Answer Accuracy: Mainly evaluates the correctness of the finally generated answer, judging whether the answer conforms to the facts and logic of the general large-scale model.
    \item System Token Consumption: It is an evaluation of the system's performance consumption.
\end{enumerate}
The value range of each of these five indicators is set within the interval [0, 1]. The final evaluation score is obtained by calculating the average value after accumulating the scores of these five indicators.

\textbf{Fidelity}: fidelity mainly measures the degree of consistency between the generated answer and the given source text (context), that is, whether the answer is accurately based on the context without deviating, exaggerating, or wrongly reflecting the information in the context. A high degree of faithfulness means that the answer given by the system can accurately rely on the provided knowledge documents (context). The calculation of faithfulness follows equation~\ref{5}, that is, it is determined by calculating the ratio of the number of words in the answer that completely match the context (NAC) to the total number of words in the answer (NA).
\begin{equation}
\text { Fidelity }=\frac{NAC}{NC}
\label{5}
\end{equation}

\textbf{Context Recall Rate}: It is used to determine whether the true answer fully appears in the recalled context. It is similar to the recall rate in classification tasks. Regardless of whether the recalled text is redundant or not, a high score can be obtained as long as the true answer can be found in it. Its score is calculated according to equation~\ref{6}, which is obtained by dividing the number of true viewpoints in the context (NTIC) by the number of viewpoints in the true answer (NTIA): 

\begin{equation}
\text { Context recall }=\frac{NTIV}{NTIA}
\label{6}
\end{equation}

\begin{figure} [h] \small
    \centering
    \includegraphics[width=0.9\linewidth]{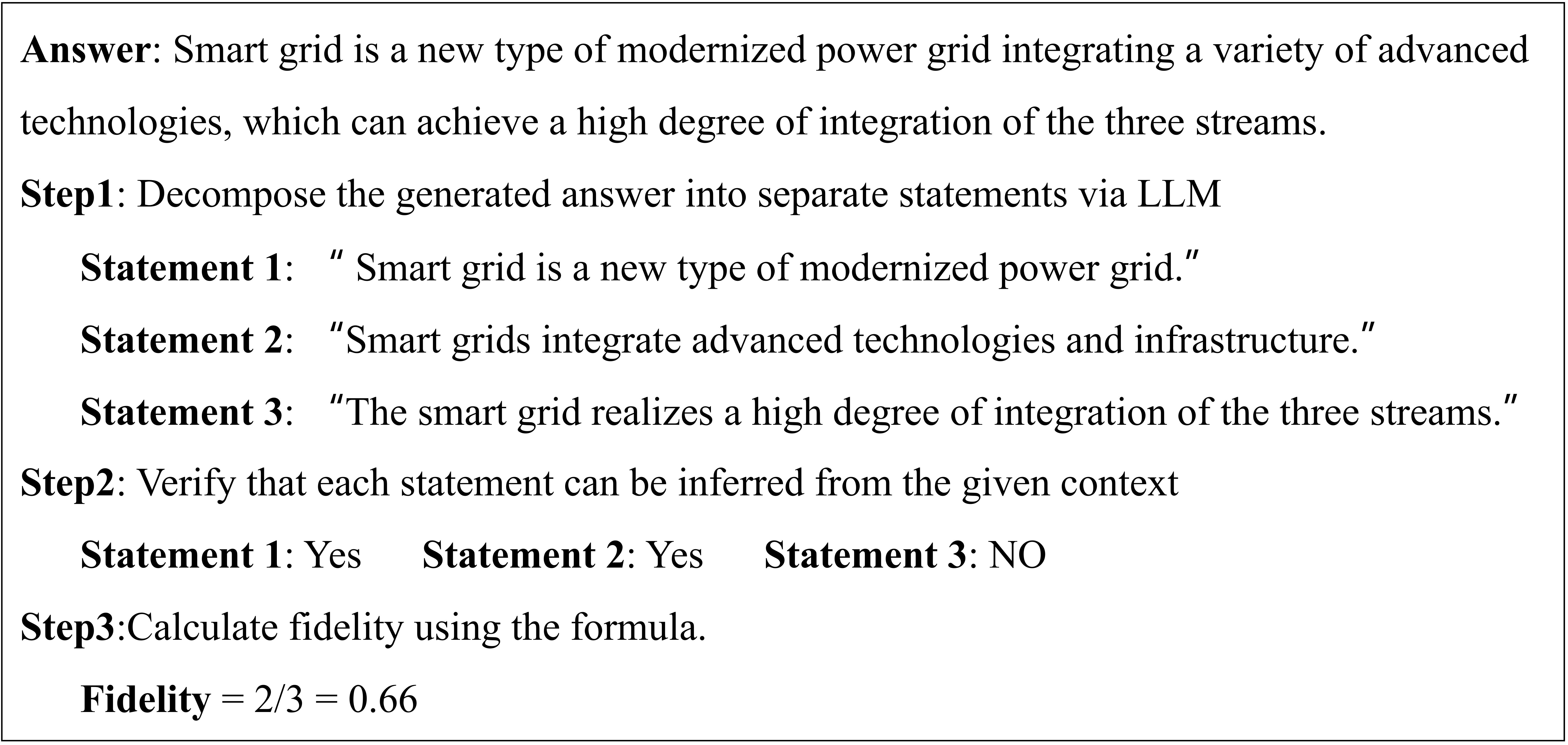}
    \vspace{-5pt}
    \caption{Contextual Recall Calculation Process}
    \label{fig:fig7}
\end{figure}

When calculating in practice, the real answer will be decomposed into several different statements, and the viewpoints supporting each statement will be searched for in the context one by one, as shown in Figure~\ref{fig:fig7}.

\textbf{Answer relevance (RA)}: RA is used to evaluate the degree of relevance between the generated answer and the original question. If the answer is incomplete or contains redundant information, it will get a lower score. If the answer directly and appropriately answers the question, it will get a higher score. Let the answer A and the question Q be represented as vectorsand,respectively. Then their cosine similarity is calculated as shown in equation~\ref{7}. When there are N expanded questions, similar calculations need to be carried out respectively and the average value is taken to obtain the cosine similarity of the question Q: 

\begin{equation}
\text Cosine\_Similarity(A,Q)=\frac{\sum_{i = 1}^{n}a_{i}b_{i}}{\sqrt{\sum_{i = 1}^{n}a_{i}^{2}}\times\sqrt{\sum_{i = 1}^{n}b_{i}^{2}}}
\label{7}
\end{equation}

\textbf{Answer accuracy} measures the accuracy of the system's answer by comparing the generated answer with the real answer. It is a weighted average score of semantic similarity and factual similarity, as shown in the formula, where $w$ represents the weight of factual relevance. The factual similarity is calculated by the F1 score. The formulas are shown in equation~\ref{8} and ~\ref{9}.

\begin{equation}
\text { Accuracy }=\frac{w * F 1+S s}{w+1}
\label{8}
\end{equation}

\begin{equation}
F 1=\frac{|T P|}{(|T P|+0.5 \times(|F P|+|F N|))}
\label{9}
\end{equation}

Where $TP$ represents viewpoints present in both the true and generated answers, $FP$ represents viewpoints present in the generated answer but not in the true answer, and $FN$ represents viewpoints present in the true answer but not in the generated answer. After obtaining the F1 score, semantic similarity is calculated using a cross-encoder-based measurement method called SAS[23].

\textbf{Token}: In the RAG system, resource consumption is usually measured by tokens. Before the text is input into the large language model, it will be converted into different numbers of tokens. Similarly, when the model outputs, it will also generate different numbers of tokens. Therefore, when using the same generation model, the total number of tokens in the input and output stages can be calculated to evaluate and compare the resource consumption of the system.

\textbf{AI Self-Check Mechanism}

In practice, large language models often produce incorrect answers because documents contain relevant context but lack sufficient context, especially when dealing with events involving specific dates, due to limitations in processing temporal information. Such errors can be circumvented by an AI self-testing mechanism [15].

\begin{figure} [h] \small
    \centering
    \includegraphics[width=0.9\linewidth]{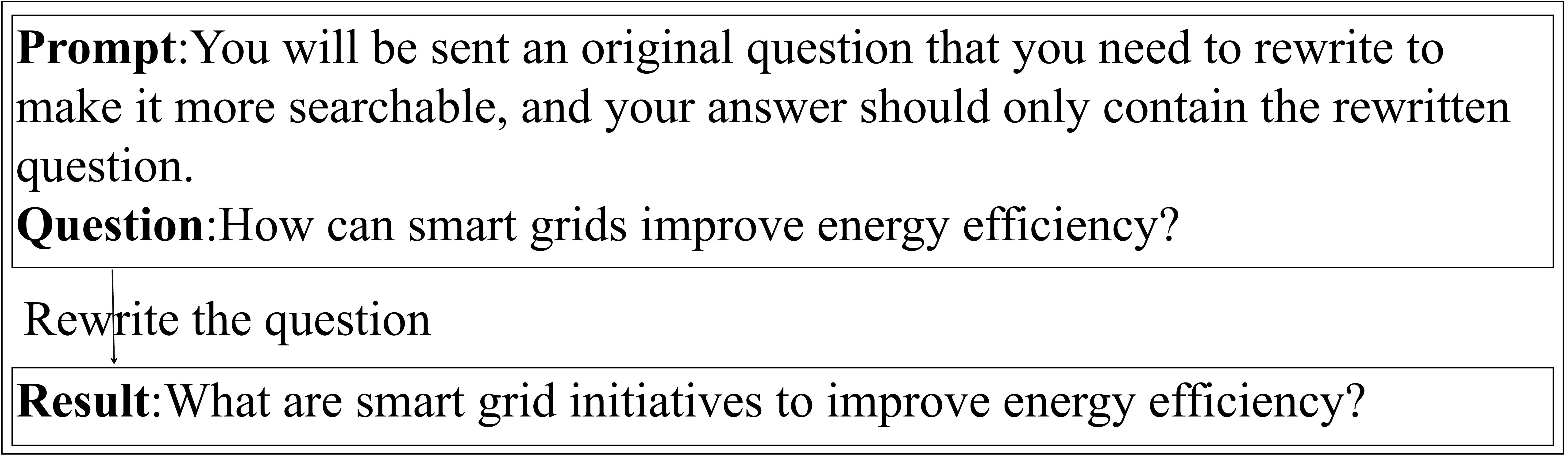}
    \vspace{-5pt}
    \caption{Example of Question Rewriting}
    \label{fig:fig8}
\end{figure}

In the AI self-checking process, we use the LLM to quantitatively score the responses based on the response quality assessment criteria provided in section \ref{bijection}. If the average score of the first four key metrics exceeds 85\%, the answer is judged as Pass; conversely, it is considered as Fail. for the Fail case, the system automatically rewrites the question and restarts the pre-retrieval phase with an iterative count increment. If the answer passes the evaluation (i.e. Pass), it will be directly output as the final answer. The system presets the maximum number of iterations to be five, beyond which if the answer still fails to meet the standard, it is considered that the system is unable to answer effectively, and it is necessary to rewrite the question according to Figure~\ref{fig:fig8} and re-enter the Q\&A system for processing.

\textbf{Algorithms of the Iterative Retrieval Q\&A Optimization Scheme}

We focuses on elaborating three core algorithms, which are introduced as follows.

The Pre-search and search stage optimization (algorithm\ref{alg:alg1}: PreRetrievalOpt) algorithm obtains the user question $Q$ through the $GetUserQ$ function and initializes it as the extended question list $ExpQList$. Then, it uses the $CallLlm$ function to call the language model to generate two extended questions for the original question and adds them to $ExpQList$. Subsequently, it uses the BM25 Retrieve algorithm of BM25 and the BGERetrieve algorithm of BGE to retrieve based on $ExpQList$ and stores the results in $CandDocs$. This dual-retrieval method combines the statistical BM25 algorithm and the semantic understanding ability of the BGE model, increasing the possibility of retrieving information relevant to the user question.
\begin{algorithm}[h!]
\label{alg:alg1}
\SetAlgoLined
\KwIn{Issues raised by users: $Q$}
\KwOut{Expanding the list of questions: $ExpOList$, Candidate documents: $CandDocs$}
\BlankLine
$Q \leftarrow$ GetUserQ()\;
$ExpO$, $CandDocs$ $\leftarrow$ PreRetrievalOpt($Q$)\;
$ExpOList$ $\leftarrow$ [$Q$]\;
\BlankLine
// Problem expansion with LLM generates two expansion problems
\For{$i \leftarrow 1$ \KwTo $2$}{
    $NewQ$ $\leftarrow$ CallLLm("Gen ext q for " + $Q$)\;
    ExpOList.Add($NewQ$)\;
}
\BlankLine
// {Combined BM25 and BGE search}

\ForEach{$q$ in ExpOList}{
    $BM25Rs$ $\leftarrow$ BM25Retrieve($q$)\;
    $BGERs$ $\leftarrow$ BGERetrieve($q$)\;
    CandDocs.Add($BM25Rs$, $BGERs$)\;
}
\BlankLine
\Return{ExpOList, CandDocs}\;
\caption{PreRetrievalOpt}
\end{algorithm}

The Post-retrieval optimization algorithm (algorithm\ref{alg:alg2}: PostRetrievalOpt) mainly optimizes the candidate documents obtained in the previous stage. First, it fine-tunes the language model through $FineTuneLlmForRating$ to obtain $PE-LLM_1$ and uses $PE1RateDoc$ to score each document in $CandDocs$. Only the documents with a score greater than 0 and their scores are stored in $FiltDocs$ in the form of tuples. Then, it uses the Sort function to sort the documents in descending order according to the ByScoreDesc standard. The final output $FiltAndSortDocs$ contains the screened and scored documents for subsequent processing.
\begin{algorithm}[h]
\label{alg:alg2}
\SetAlgoLined
\KwIn{Candidate documents: $CandDocs$}
\KwOut{Filtered, scored documents: $FiltAndSortDocs$}
\BlankLine
FiltAndSortDocs $\leftarrow$ PostRetrievalOpt($CandDocs$)\;
\BlankLine
// {Fine-tuning LLM gets $PE-LLM_1$}

 $PE-LLM_1$ $\leftarrow$ FineTuneLlmForRating()\;
\BlankLine
// {Filtering, scoring and sorting candidate documents}

FiltDocs $\leftarrow$ []\;
\For{$doc$ In $CandDocs$}{
    \If{score $\leftarrow$ PERateDoc(doc) > 0}{
        FiltDocs.Add($doc$, $score$)\;
    }
}
\BlankLine
SortDocs $\leftarrow$ Sort($FiltDocs$, $ByScoreDesc$)\;
\Return{FiltAndSortDocs}\;
\caption{ PostRetrievalOpt}
\end{algorithm}

The Generation phase optimization algorithm (algorithm\ref{alg:alg3}: GenOpt) is responsible for generating the final answer to the user question. It takes the extended question list $ExpQList$ and the screened and scored documents $FiltAndSortDocs$ as input and uses the $CallLlmToGenAns$ function to generate the initial answer. Then, it fine-tunes the language model through $FineTuneLlmForAnsEval$ to obtain $PE-LLM_2$ and uses $PE2EvalAns$ to evaluate the generated answer $GenAns$. If the evaluation result $EvalRes$ does not meet the requirements and the iteration count $IterCount$ is less than 5, it rewrites the question using $RewriteQ$ based on the current answer, performs pre-retrieval and post-retrieval optimization again, and generates a new answer. The iteration continues until the evaluation passes or the maximum iteration count is reached. If the evaluation passes, the generated answer is the final answer $Ans$. The iteration process in this stage uses the feedback of the evaluation language model to continuously optimize the answer and improve the answer quality.
\begin{algorithm}[h]
\label{alg:alg3}
\SetAlgoLined
\KwIn{Expanding the list of questions: $ExpOList$; Filtered, scored documents: $FiltAndSortDocs$}
\KwOut{Results of the question and answer session: $Ans$}
\BlankLine
$Ans$ $\leftarrow$ GenOpt($ExpOList$, $FiltAndSortDocs$)\;
\BlankLine
// {Calling LLM to Generate Answers}

$GenAns$ $\leftarrow$ CallLlmToGenAns($ExpO$, $FiltAndSortDocs$)\;
\BlankLine
// {Fine-tuning LLM to get $PE-LLM_2$ assessment answers and self-tests}

$PE-LLM_2$ $\leftarrow$ FineTuneLlmForEval()\;
\BlankLine
$EvalRes$ $\leftarrow$ PE2EvalAns($GenAns$)\;
\BlankLine
IterCount $\leftarrow$ 0\;
\If{If the AI self-test fails, rewrite the question and re-enter the process for up to 5 iterations.}{
    \While{Not EvalRes and IterCount < 5}{
        $RewrittenQ$ $\leftarrow$ RewriteQ($GenAns$)\;
        $NewExpO$, $NewCandDocs$ $\leftarrow$ PreRetrievalOpt($RewrittenQ$)\;
        $NewFiltAndSortDocs$ $\leftarrow$ PostRetrievalOpt($NewCandDocs$)\;
        $GenAns$ $\leftarrow$ CallLlmToGenAns($NewExpO$, $NewFiltAndSortDocs$)\;
        $EvalRes$ $\leftarrow$ PE2EvalAns($GenAns$)\;
        $IterCount$ $\leftarrow$ $IterCount$ + 1\;
    }
}
\Return{$Ans$}\;
\caption{GenOpt}
\end{algorithm}

%% file: sections/07_evaluation.tex
\section{Case Study}
In this chapter, we first introduce the dataset. Then, we introduce the experimental environment, and the project code has been made public. The access address is: {https://gitee.com/zhangjiqun/qa\_system}. Finally, we focus on presenting the experimental results. This experimental part is divided into three parts: the first part is the retrieval method comparison experiment, which aims to verify the superiority of the combination of sparse retrieval and dense retrieval strategies proposed in this paper; the second part is to conduct a system ablation experiment based on different answer evaluation indicators to evaluate the specific contribution of each indicator to the system performance; the third part is the overall system comparison experiment, which aims to comprehensively compare and display the overall optimization effect achieved by this scheme.
\subsection{Dataset Construction}

Currently, mainstream question answering datasets, such as Triviaqa [24] and Natural Questions, mostly collect data from Wikipedia. The characteristics of this data source are likely to cause test bias in pre-trained large language models, and general large models have disadvantages in adapting to the smart grid question answering system and are difficult to meet the needs of this specific field.
In view of this, this study selects the dataset involved in reference [22] and makes some data corrections on its existing basis, and then constructs the SGQA (Smart grid question and answer) dataset. This dataset has been made public, and the detailed website is: https://gitee.com/zhangjiqun/sgqa.
The SGQA dataset is a dataset specially designed for knowledge question answering tasks in the power field. It contains 33,500 power specification clauses and 20,000 question and answer pairs. The power specification clauses focus on training the LLM to learn the basic theoretical knowledge of power specifications and help the model master the basic principles of the power field. The question and answer pairs focus on the in-depth learning of power-related knowledge points, and the question and answer pairs can also provide a practical reference for model evaluation to ensure that the performance of the model in the power knowledge question answering scenario can be accurately measured.
The SGQA dataset covers a wide range of fields, comprehensively covering various key knowledge points in the power field such as thermal power technology, hydropower station equipment maintenance management, power capacitor and inductor testing, electrical materials, power construction, wind farm power assessment, power transformer selection, nuclear power plant equipment maintenance, power plant boiler unit dust collector maintenance, phase modifier maintenance, and overhead transmission line high-altitude rescue. From the perspective of question type distribution, the dataset includes three mainstream question types: single choice, fill in the blanks, and judgment. Table~\ref{tab:tab2} lists in detail the number and answer content of each question type in the dataset, and Table~\ref{tab:tab3}  shows sample examples of each question type. The "reference" field in Table 3 clearly marks the source of the question, providing convenience for users to understand the background of the question and trace the knowledge context.

\begin{table}[htbp]
\centering
\caption{Distribution of questions by category}
\begin{tabular}{lll}
\toprule
Question Category & Number & Answer content \\
\midrule
Single choice & 6000 & A|B|C|D \\
judgment & 7000 & - \\
fill in the blanks & 7000 & True/False \\
\bottomrule
\end{tabular}
\label{tab:tab2}
\end{table}

\begin{table}[htbp]
\centering
\caption{Sample examples of each type of question}
\begin{tabular}{p{0.04\textwidth}p{0.18\textwidth}p{0.05\textwidth}p{0.16\textwidth}}
\toprule
Category& Question & Answer & Reference \\
\midrule
Single choice & Which of the following terms and definitions is not applicable to this document? A: GB/T 2900.5 B: GB/T 25096 C: GB/T 8287.1 D: ISO 9001:2015 & D & The terms and definitions defined in GB/T 2900.5, GB/T 2900.8, GB/T 8287.1 and GB/T 25096 are applicable to this document. \\
 
judgment & The area of a single surface defect of a composite insulator does not exceed 25.0mm², the depth does not exceed 1.0mm, the protrusion height does not exceed 0.8mm, and the mold joint is flat & True & b) Composite insulator The area of a single surface defect does not exceed 25.0mm², the depth does not exceed 1.0mm, the protrusion height does not exceed 0.8mm, and the mold joint is flat. \\
 
fill in the blanks & According to the requirements of 7.2, on what facilities should the unpacked insulators be placed to protect the umbrella cover? Please fill in the keywords. & Protection measures & The unpacked insulators should have protection measures to avoid deformation or damage of the umbrella cover. \\
\bottomrule
\end{tabular}
\label{tab:tab3}
\end{table}

\textbf{Benchmark and Datasets}  
We utilize four commonly employed real-world datasets.

\subsection{Experimental Environment} %
The experimental environment configuration of this paper is shown in Table\ref{tab:tab4}. The experiment is carried out based on the Ubuntu 22.04 system. The computer hardware configuration used is: equipped with 5 NVIDIA A100-PCIE-40GB GPUs, each GPU has 6912 CUDA cores, with a total video memory capacity of 200GB and a frequency of 3.2GHz; the CPU adopts Intel® Xeon® Gold 6248R, with a main frequency of 3.00GHz and 96 cores; the memory is 1007.3GB, and the disk capacity is 7.8TB.

\begin{table}
\caption{Experimental Environment}
\centering
\begin{tabular}{p{0.15\textwidth}p{0.23\textwidth}}
\toprule[0.8pt]
Hardware/Environment & Specification \\
\hline
Operating system & Ubuntu 22.04 \\
GPU & 5 * NVIDIA A100-PCIE-40GB \\
CUDA cores & 6912 \\
Video memory & 200GB \\
memory frequency & 3.2GHz \\
CPU & Intel® Xeon® Gold 6248R CPU @ 3.00GHz * 96 \\
Memory & 1007.3 GB \\
Disk capacity & 7.8TB \\
Python & 3.10.13 \\
PyTorch & 2.2.1 \\
CUDA & 11.8 \\
Ragas & 0.1.2 \\
Llama-index & 0.10.12 \\
\bottomrule[0.8pt]
\end{tabular}
\label{tab:tab4}
\end{table}

At the software level, the Python version is 3.10.13. PyTorch 2.2.1 is a deep learning framework used to build and train neural network models. CUDA 11.8, as a parallel computing platform and programming model launched by NVIDIA, enables GPUs to perform general computing and cooperates with PyTorch to improve the training speed of deep learning.

In addition, two important toolkits are used. Ragas 0.1.2 is mainly used to evaluate applications based on language models and help optimize model performance. Llama-index 0.10.12 realizes the management of large language models and improves development efficiency.

\subsection{Experimental Results}
\textbf{Retrieval Model Comparison Experiment}

To verify the effectiveness of the retrieval scheme proposed in this paper, we compares it with several common retrieval models, conducting multiple experiments to measure their performance differences.

\textbf{BM25 Algorithm~\cite{robertson2009probabilistic}:} It is a classic algorithm for information retrieval and one of the representative algorithms for sparse retrieval, commonly used in search engines. It assesses the relevance score between documents and queries based on the frequency of query terms in the documents and ranks them accordingly.

\textbf{BGE Model~\cite{luo2024bge}:} The bge-base-zh-v1.5 model from the series, developed by the Beijing Academy of Artificial Intelligence, is used in this experiment. It is an advanced dense retrieval method that converts text into dense embedding vectors and calculates the cosine similarity between different text vectors to find similar texts.

\textbf{Method Used in This Paper:} This paper enhances retrieval recall quality and robustness by using a parallel approach of sparse retrieva and dense retrieval. Since this method retrieves twice the number of texts as a single retriever, LLM is employed to score and filter the retrieved documents to obtain the same number of candidate documents as other schemes.


\begin{figure} [h] 
    \centering
    \includegraphics[width=0.9\linewidth]{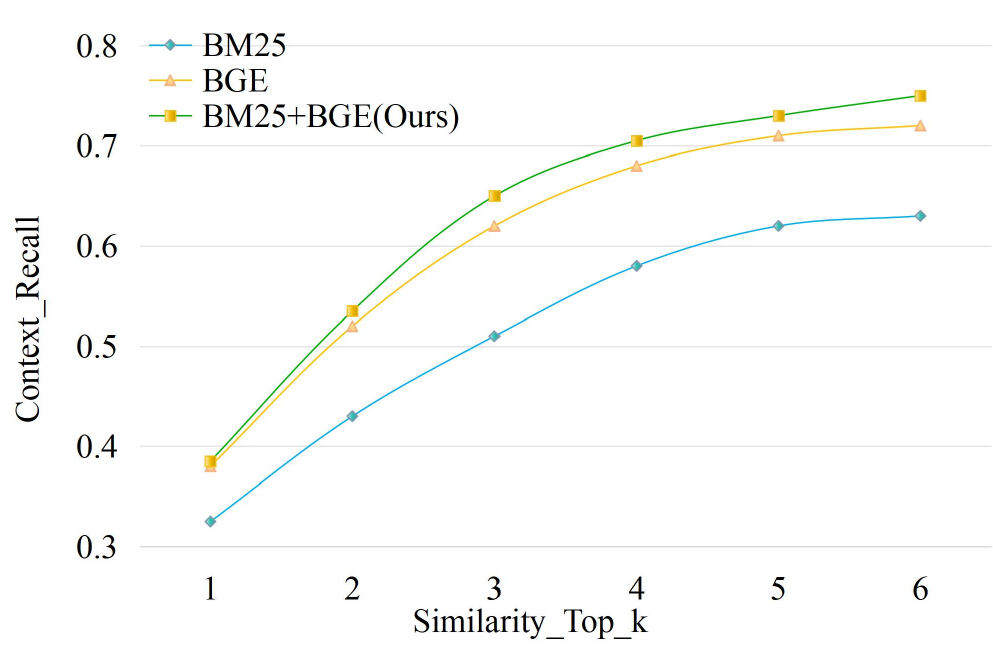}
    \caption{Comparison of Retrieval Methods in Experiments}
    \label{fig:fig9}
\end{figure}

Since too many query expansions will cause the system to spend too much, and too few expansions may not significantly improve the performance, in order to find the optimal number of query expansions (NIE), this paper conducts an experiment on the NIE when using the hybrid retrieval and fixing the recall text number to 5. It can be seen from Table~\ref{tab:tab5} that when the number of queries is expanded to 2, the growth rate of the recall rate is less than 1\%, and the trend becomes flat. In the actual question generation process, too many query expansion operations may lead to serious homogenization of the generated queries, which will greatly increase the system consumption while having a small improvement effect on the system answering effect. Therefore, finally, the number of query expansions in the system is set to 2.

\begin{table}[htbp]
\centering
\caption{Effect of the number of query expansion on contextual recall}
\begin{tabular}{cccccc}
\toprule
NIE & 0 & 1 & 2 & 3 & 4 \\
\midrule
Context\_Recall & 0.74 & 0.785 & 0.82 & 0.825 & 0.828 \\
Growth\_Rate & 6.08\% & 4.46\% & 0.61\% & 0.36\% & - \\
\bottomrule
\end{tabular}
\label{tab:tab5}
\end{table}

\textbf{System Ablation Experiment}

In this ablation experiment, this paper selects a naive RAG system as the baseline and consistently uses the Gpt3.5-Turbo model as the generator and the bge-base-zh-v1.5 model as the embedding vector model, with the maximum number of iterative retrievals set to three. The overall system optimization is divided into three parts, corresponding to different stages of the RAG system: pre-retrieval and retrieval optimization (query expansion + hybrid retrieval), retrieval and post-retrieval optimization (document filtering + reordering), and iterative retrieval (enabling the model to perform a new round of retrieval through answer self-checking and rewriting). The Ragas framework is used to score four indicators: fidelity, context recall rate, answer relevance, and answer accuracy.

The experimental results in Table \ref{tab:tab6} show that the baseline system performs poorly in context recall rate, which in turn negatively impacts answer accuracy due to its lack of optimization and reliance on a single retrieval method. By incorporating query expansion and hybrid retrieval strategies, the system’s retrieval scope is expanded, significantly improving the baseline model’s context recall rate and indirectly enhancing fidelity, answer relevance, and answer accuracy. The further introduction of document filtering and a redesigned reordering mechanism effectively improves context recall quality by eliminating irrelevant candidate documents, while substantially enhancing fidelity and answer accuracy. Finally, the introduction of an answer self-checking mechanism enables iterative improvements in the answers, further enhancing fidelity and answer accuracy. After these three optimizations, the RAG system described in this paper outperforms the baseline system by 15.69\% in fidelity, 22.64\% in context recall rate, 8.32\% in answer relevance, and 22.25\% in answer accuracy.

\begin{table}[h]\small
 \caption{Ablation Experiment Results}
    \centering
     \vspace{-5pt}
    \scalebox{0.55}{
   \begin{tabular}{ccccccc}
\toprule[1.5pt]
\begin{tabular}{l} 
Query \\
Expansion + \\
Hybrid Retrieval
\end{tabular} & \begin{tabular}{l} 
Document \\
Filtering + \\
Reordering
\end{tabular} & \begin{tabular}{l} 
Iterative \\
Retrieval
\end{tabular} & fidelity (\%) & \begin{tabular}{l} 
Context \\
Recall \\
Rate(\%)
\end{tabular} & \begin{tabular}{l} 
Answer \\
Relevance(\%)
\end{tabular} & \begin{tabular}{l} 
Answer \\
Accuracy(\%)
\end{tabular} \\
\hline$\times$ & $\times$ & $\times$ & 78.85 & 59.73 & 86.49 & 56.46 \\
$\sqrt{ }$ & $\times$ & $\times$ & 84.32 & 69.87 & 89.82 & 67.49 \\
$\sqrt{ }$ & $\sqrt{ }$ & $\times$ & 91.38 & 82.25 & 94.27 & 74.76 \\
$\sqrt{ }$ & $\sqrt{ }$ & $\sqrt{}$ & 94.54 & 82.37 & 94.81 & 78.71 \\
\bottomrule[1.5pt]
\end{tabular}}
\label{tab:tab6}
\end{table}

\textbf{System Comparison Experiment}

To verify the effectiveness of the optimization scheme, this study conducts comparative experiments with several other RAG Q\&A systems. The control schemes include: Naive RAG, RAG optimized using the HyDE method, ITRG, and Self-RAG, which are described as follows:

\textbf{Naive RAG~\cite{gao2023retrieval}:} A system built from the most basic RAG process.

\textbf{HyDE~\cite{gao2022precise}:} This method generates pseudo-answers using an LLM and uses these pseudo-answers as queries to retrieve real texts, thereby improving the system's context recall rate. Compared to directly using the original question for retrieval, these pseudo-answers are likely to be closer to the real text in the vector space, thereby enhancing the relevance and accuracy of retrieval.

\textbf{ITRG~\cite{feng2024retrieval}:} It can be viewed as an iterative retrieval version of the HyDE scheme. The core idea is that each retrieved text contains only partial information to answer the question. Therefore, the answer can be progressively refined through repeated generation of incomplete answers and retrieval of real texts based on these incomplete answers. In each iteration, the system generates an answer that may contain both false and partially correct information, which is then used as a query to retrieve similar texts, thereby collecting additional information to refine the answer.

\textbf{Self-RAG~\cite{asai2023self}:} This method introduces retrieval and evaluation tokens to the model. The retrieval token is used to allow the LLM to determine whether the retrieval process should be executed for a given question, while the evaluation token is employed for document filtering and ranking. Adaptive and iterative retrieval are facilitated through these tokens.

\begin{figure} [h] 
    \centering
    \includegraphics[width=0.9\linewidth]{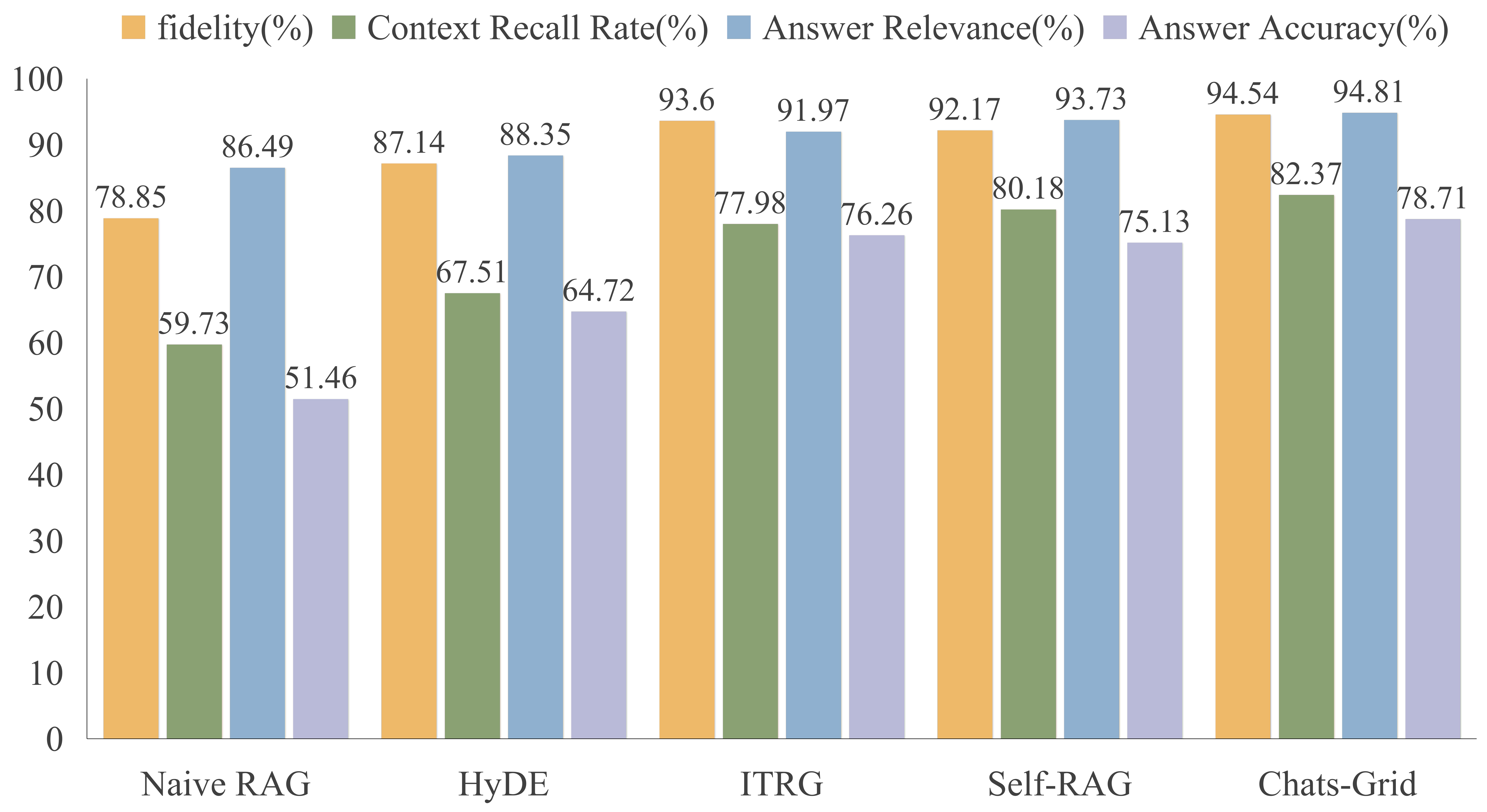}
    \caption{Comparison of Retrieval Methods in Experiments}
    \label{fig:fig10}
\end{figure}

\begin{table}[htbp]
\centering
\caption{Comparison of different RAG systems}
\begin{tabular}{ccccccc}
\toprule
RAG System & Naive RAG & HyDE & ITRG & Self-RAG & Ours \\
\midrule
Cost (tokens) & 397 & 464 & 703 & 769 & 826 \\
\bottomrule
\end{tabular}
\label{tab:tab8}
\end{table}

As can be seen from the data in Figure~\ref{fig:fig10}, the plain RAG underperforms in all aspects due to the fact that no additional optimisations have been implemented. However, it is this unoptimised nature that allows the method to consume relatively few tokens, and HyDE effectively improves answering by optimising the quality of retrieval. Although this optimisation strategy leads to improved response results, it comes at the cost of a slight increase in system overhead. both ITRG and Self-RAG use an iterative approach to achieve significant performance improvements in all aspects of the Q\&A system. However, the iterative process inevitably leads to a further increase in token consumption, as shown in Table~\ref{tab:tab8}.

The experimental results show that our proposed improvement scheme achieves better results in the four key metrics of fidelity, contextual recall, answer relevance, and answer accuracy. Specifically, in the Q\&A test with 100 pieces of data, the average system overhead of the proposed scheme increases by only 57 tokens per Q\&A session compared to the Self-RAG scheme, which confirms that our improved scheme achieves a good balance between performance improvement and system overhead control, and provides a more advantageous solution for the Phase-Intelligent Grid Q\&A system.

%% file: sections/10_conclusion.tex
\section{Conclusion}

This paper systematically presents a comprehensive analysis of the traditional construction process for Q\&A systems, detailing key stages such as indexing strategies, retrieval techniques, and model selection. We introduce Chat-Grid, a cost-effective solution aimed at raising efficiency and answer quality in smart grid. In the pre-retrieval and retrieval stages, Chats-grid utilize a combination of dense and sparse retrievers to maximize the efficiency and coverage of the document retrieval process. In the post-retrieval stage, we employ a LLM to assess the relevance of retrieved documents, filtering out irrelevant content and re-ranking the results to enhance retrieval accuracy. Furthermore, we propose an innovative model self-checking mechanism coupled with question reformulation, enabling iterative retrieval that ensures the consistency and correctness of the answers by identifying and addressing inconsistencies in the facts retrieved.

Through three sets of experiments: retriever comparison, ablation study, and system comparison. We demonstrated the significant improvements brought by our proposed Chats-Grid scheme. The experimental results indicate that our approach significantly outperforms existing methods such as Self-RAG and ITRG in terms of fidelity, contextual recall rate, and answer accuracy. Specifically, Chats-Grid shows improvements of 2.37\%, 2.19\%, and 3.58\% in fidelity, context recall rate, and answer accuracy over Self-RAG, respectively, and 0.94\%, 4.39\%, and 2.45\% over ITRG. These findings confirm the effectiveness and potential of the proposed optimization strategy.

In future research, we will focus on the optimization of computational efficiency and expanding the implementation of the proposed system in real-world smart grid and other domain-specific applications to evaluate its practical effectiveness. Further refinement of the iterative retrieval process to reduce the computational overhead while maintaining high retrieval accuracy. We are looking forward to develop more advanced self-checking algorithms that improve system accuracy without significantly increasing complexity. 

%